\pdfoutput=1

\documentclass[11pt]{article}

\usepackage{acl}

\usepackage{times}
\usepackage{latexsym}

\usepackage[T1]{fontenc}

\usepackage[utf8]{inputenc}

\usepackage{microtype}

\usepackage{inconsolata}

\newcommand{\methodName}{\textsc{Epicure}\xspace} 

\usepackage[paperVersion]{dpupaper}

\usepackage{etoolbox}
\newcommand{\kptdisplayskips}{%
  \setlength{\abovedisplayskip}{4pt}%
  \setlength{\belowdisplayskip}{4pt}%
  \setlength{\abovedisplayshortskip}{4pt}%
  \setlength{\belowdisplayshortskip}{4pt}}
\appto{\normalsize}{\kptdisplayskips}
\appto{\small}{\kptdisplayskips}
\appto{\footnotesize}{\kptdisplayskips}

\title{\methodName: Distilling Sequence Model Predictions into Patterns}

\author{Miltiadis Allamanis\thanks{~~Work performed while the author was unemployed. 
        Now at Google DeepMind. \texttt{mallamanis@google.com} } \\
  \\\And
  Earl T. Barr \\
  UCL, United Kingdom \\
  \texttt{e.barr@ucl.ac.uk} \\}

\begin{document}
\maketitle
\begin{abstract}

Most machine learning models predict a probability distribution over concrete outputs and struggle to accurately predict names over high entropy sequence distributions.  Here, we explore finding abstract, high-precision patterns intrinsic to these predictions in order to make abstract predictions that usefully capture rare sequences.  In this short paper, we present \methodName, a method that distils the predictions of a sequence model, such as the output of beam search, into simple patterns.  \methodName maps a model's predictions into a lattice that represents increasingly more general patterns that subsume the concrete model predictions.

On the tasks of predicting a descriptive name of a function given the source code of its body and detecting anomalous names given a function, we show that \methodName yields accurate naming patterns that match the ground truth more often compared to just the highest probability model prediction.  For a false alarm rate of 10\%, \methodName predicts patterns that match 61\% more ground-truth names compared to the best model prediction, making \methodName well-suited for scenarios that require high precision.
\end{abstract}

\section{Introduction}

Machine learning models that predict sequences commonly yield a probability distribution over concrete outputs.
High entropy distributions (many plausible predictions with low probability) are common in applications of sequence models, especially when you condition on the fact that models are rarely queried about obvious names.
In this short paper, we present \methodName,\footnote{We call our approach \methodName, because it is a discerning epicure that discovers the ingredients in a ``dish'' of sequences of concrete predictions.} a method that distils a number of probabilistic predictions into high-level, abstract patterns (\autoref{sec:method}).
For example, \methodName may distil a set of prediction to the abstract pattern \texttt{A$\star$C} and assign it a probability of 60\%, which means that the model providing the predictions predicts a 60\% chance that the output should start with \texttt{A} and end with \texttt{C} with zero or more arbitrary elements appearing in-between.
Such patterns can be useful for making (partial) suggestions that contain the ``essence'' of the prediction but eschew hard-to-predict or ambiguous parts of the suggested sequence.
Such patterns are also useful for interpretable anomaly (anti-pattern) detection, where false alarms need to be rare.

We evaluate \methodName on the function naming and anomalous function name detection tasks (\autoref{sec:eval}), \ie the task of predicting a descriptive name of a function given the source code of its body or detecting that an exciting name is anomalous.
These tasks are a form of ``extreme'' summarisation~\citep{allamanis2016convolutional}.

\section{Background and Related Work}

Commonly, NLP and machine learning models aim to predict a concrete output rather than make a partial/abstract prediction.
\citet{weiss2019learning} presents a method that learns a weighted automaton from an unconditional sequence model to study the
learning capacity of various sequence models.
Closer to our work is Grammformers~\citep{guo2021learning} that use reinforcement learning, grammar-driven code generation, to generate sketches, \ie code utterances with holes which can be thought as patterns.
In contrast, \methodName does not require a grammar --- which may not exist --- and is simpler and computationally more efficient, requiring only the outputs of a sequence model with \emph{no} additional training.

\paragraph{Naming Patterns in Code} 
Inconsistent, misleading, or incorrect naming in code has been shown to negatively affect code maintenance, \eg readability~\citep{lawrie2007effective}.
Detecting bad/anomalous names~\citep{host2009debugging} and ``linguistic anti-patterns''~\citep{arnaoudova2016linguistic} has focused on hand-written rules and heuristics.
\citet{arnaoudova2016linguistic} manually create a taxonomy of such anti-patterns in Java, \eg ``\emph{method starts with \texttt{is} but returns more than a Boolean}'', and ``\emph{Set method returns}'' a value.
When used for anomaly detection, \methodName automates detecting inconsistent, misleading, or incorrect names.

\paragraph{Predicting Names with Machine Learning}
The idea of predicting meaningful identifier names with machine learning originates from the concept of code's naturalness~\citep{hindle2012naturalness}, \ie that source code shares many characteristics with natural languages.
\citet{allamanis2014learning} first showed that language models of code can predict local variable names.
This results has been replicated --- and improved --- by multiple research works, such as the log-bilinear model of \citet{allamanis2015suggesting}, the distributed representations of \citet{bavishi2018context2name}, and the graph-based representations of \citet{allamanis2018learning}.
In contrast to \citet{allamanis2014learning}, this works predict variable names as a sequence of subtokens rather than single units.
Predicting names as a sequence of subtokens affords greater flexibility in generating novel names (``neologisms'') and takes into account the diverse nature of identifiers in source code~\citep{allamanis2013mining}.
The same idea has been researched for code deobfuscation~\citep{raychev2015predicting,vasilescu2017recovering,lacomis2019dire,tran2019recovering} --- a harder task requiring to jointly predict the names of multiple entities, such as local variables and parameters that interrelate.

Learning to name functions is commonly harder than naming local variables because it requires a holistic view of the function code; it has been a research focus.
\citet{allamanis2015suggesting} first used a log-bilinear model to suggest method names as a sequence of subtokens, and
\citet{bichsel2016statistical} deobfuscated method names in Android applications.
Later, \citet{allamanis2016convolutional} used an attention-based deep learning model to predict function names as a sequence of subtokens.
More recently, \citet{alon2018general,alon2018code2seq} improved the state-of-the-art by considering leaf-to-leaf paths over code's syntax tree.
With the advent of transformers~\citep{vaswani2017attention}, encoder-decoder and decoder-only models have been used to predict function names and natural language summaries.

All this work is orthogonal to \methodName, which distils the predictions of any of these models into interpretable patterns to suggest names or detect anomalous names.

\section{\methodName: Extracting Ingredients}\label{sec:method}

We present \methodName, a method for distilling a set of ``concrete''
predictions into a more probable abstract, yet interpretable, pattern that captures the ingredients
of the concrete set.  Consider a vocabulary of
subtokens $V$.  Let $T=\{t\}$ where $t=[t_1, t_2, ..., t_N]\in
V^*$, $t_i \in V^+$, be a set of high-probability predictions.  A machine learning model assigns a probability to
each $t\in T$; it maps $P:T\rightarrow (0,1]$, and $\sum_{t\in T} P(t)
\leq 1$.
Most sequence-generating machine learning methods yield $T$ and $P$ through beam search or sampling.
A pattern $s$ is a standard regular expression: a sequence of tokens, including any concrete $t$, and special symbols.

\begin{figure}[t]
  \includegraphics[width=\columnwidth]{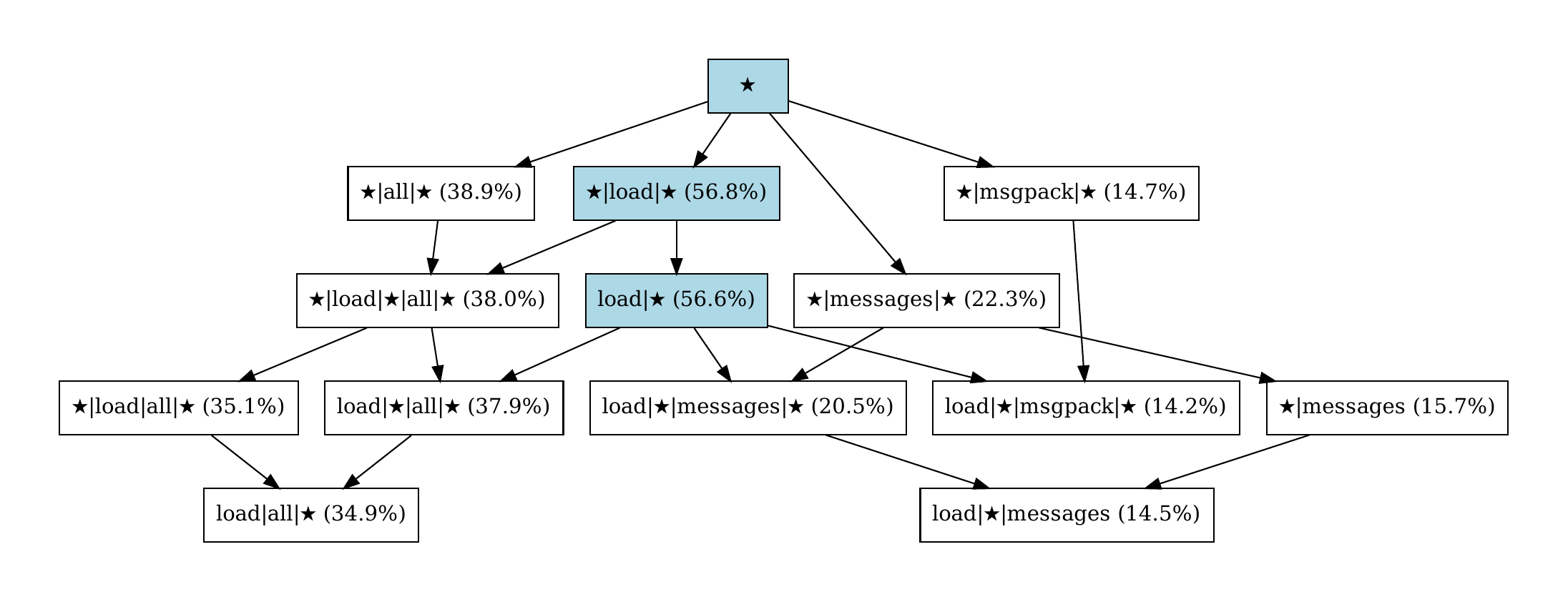}
  \vspace*{-6mm}
  \caption{A lattice for a function with ground-truth name \texttt{load\_all\_msgpack\_l\_gz}.
  Each node represents the pattern $s$ of the elements it matches with its subtokens delimited by ``|''.
  Only nodes with probability more than 13\% are shown for visual clarity.
  The nodes above a threshold $\theta$ of 55\% are shaded blue.
  \autoref{fig:lattice sample big} (\autoref{sec:lattice example}) shows more lattice nodes.
  }
  \vspace*{-4mm}
  \label{fig:lattice sample}
\end{figure}

\paragraph{Definitions}

For the pattern $s$ and a sequence $t$, let $m(s,t)$ indicate whether 
$s$ matches $t$. A pattern $s_i$ \emph{subsumes} (is more general
than) $s_j$, written
\begin{align*}
	s_j \sqsubseteq s_i \text{ iff } 
	\{t \mid m(s_j,t) \} 
		\subseteq \{t \mid m(s_i, t) \}.  
\end{align*}

\methodName's goal is to construct a lattice, like the one shown in \autoref{fig:lattice sample}.
To built this lattice, we define the
join operation $s_m = s_1 \sqcup s_2$ that accepts
two patterns and yields the \emph{least general pattern} $s_m$ that matches
both $s_1$ and $s_2$, \ie, formally $s_i \sqsubseteq
s_m$ and $s_j \sqsubseteq s_m$ and $\nexists s_k: s_k \sqsubset s_m
\wedge s_i \sqsubseteq s_k \wedge s_j \sqsubseteq s_k$. 

In this work, the join operation merges two patterns by replacing different
segments with $V^*$, \ie zero or more
arbitrary subtokens in $V$.  For brevity, we introduce the wildcard $\star = V^*$.
For example, 
\begin{align*}
	[A, B, C]  &\sqcup  [A, D, C] &&= [A, \star, C]\\
	[A, B, C, D]  &\sqcup  [A, E, B] &&= [A, \star, B, \star]\\
	[A, \star, B, \star]  &\sqcup [A, \star, C] &&= [A, \star]
\end{align*}
We chose this implementation of join because it yields interpretable patterns at the cost of over-generalising.
This implementation is a form of anti-unification~\citep{pfenning1991unification}.
Future work may profitably explore defining join to produce less restrictive 
regular expressions, albeit at the cost of problematising the connection to anti-unification.

To compute $s_1 \sqcup s_2$, we find their longest, non-overlapping subsequences\footnote{We use \texttt{SequenceMatcher} and its algorithm from Python's \texttt{difflib} standard library, which is based on ``gestalt pattern matching''~\citep{ratcliff1988gestalt}.}.
Then, $s_m$ is computed by processing each matching block.
Matching blocks are retained.
Every block that was added, deleted, or replaced between $s_1$ and $s_2$ is replaced with a wildcard.
Finally, contiguous wildcards ($\star$) are replaced with a single one.

\begin{algorithm}[t]
  \caption{Construct Lattice}\label{alg:lattice}
\begin{algorithmic}
  \State $R \gets T$ \Comment{Initialise with concrete suggestions.}
  \State $c\gets\{\}$ \Comment{Initialise parent to children map.}
  \While{$|R| > 1$}
  \State $R'\gets\{\}$
  \ForAll{$s_i, s_j\in R, s_i\neq s_j$}
    \State $s_m \gets s_i \sqcup s_j$
    \State $R'\gets R'\cup\{s_m\}$
    \If{$s_i\neq s_m$}
    \State $c[s_m]\gets \left(c[s_m]\cup\{s_i\}\right)/c[s_i]$
    \EndIf
    \If{$s_j\neq s_m$}
    \State $c[s_m]\gets \left(c[s_m]\cup\{s_j\}\right)/c[s_j]$
    \EndIf
  \EndFor
  \State $R\gets R'$
  \EndWhile
	\Return $R$ 
\end{algorithmic}
\end{algorithm}

\paragraph{Lattice}
Given the join operator ($\sqcup$), we create a lattice using \autoref{alg:lattice} (\autoref{fig:lattice sample} shows an example) that summarises the predictions of a model into high-level patterns in its outputs.
All the singleton sets of the lattice are concrete suggestions in $T$; $P$
assigns a probability to each one.
To compute the probability of a pattern $s$, we sum the probabilities of the concrete suggestions in $T$ it matches:
\begin{align}\label{eq:prob sum}
	P(s) = \sum_{t\in T, m(s,t)} P(t),
\end{align}
which is a lower bound on the probability the model assigns to all suggestions matching $s$.

\paragraph{Using the Lattice}
Given a lattice and a probability threshold $\theta \in (0.5, 1]$, we select the set of the
\emph{least general patterns} (lattice elements) $\{s^*\}$ that exceed the threshold, \ie $P(s^*) > \theta$ and $\nexists s' \sqsubset s^*:P(s')>\theta$.
Since the lattice defines a partial ordering, it may contain several incomparable elements (\ie $s_i, s_j: s_i {\not\sqsubseteq} s_j \vee s_j {\not\sqsubseteq} s_i$) above the threshold; all of them are returned.  For example, setting the threshold to 20\% in \autoref{fig:lattice sample} returns $\texttt{load}|\texttt{all}|\star$ and $\texttt{load}|\star|\texttt{messages}|\star$.

If the task is to suggest a naming pattern, \methodName yields $\{s^*\}$.
For anomaly detection, we expect that a good identifier matches \emph{all} patterns in $\{s^*\}$.
If not, we deem the name anomalous and raise an alarm. 
Two patterns $\{s^*\}$ are incompatible when their token sets are disjoint.
To eliminate incompatible patterns, we require $\theta > 0.5$.
Since $T$ is the output of a probabilistic model, 
each pattern matches at least two concrete suggestions in $T$, making it impossible for two incompatible patterns to have a probability
greater than 50\%, since they would have to match at least one common $t\in T$.

\paragraph{Limitations} \methodName is limited by the upstream model yielding $T$. For example, if the beam search that yields $T$ gets stuck to a local optimum, then \methodName cannot provide a reasonable distillaton.

\section{Evaluation}
\label{sec:eval}
We evaluate \methodName on naming functions and detecting anomalous funciton names in source code.
Despite this choice, \methodName is more broadly applicable to other identifiers such as formal parameters, local variables, \etc as well as arbitrary outputs of sequence models, \eg in machine translation and image captioning. We leave these to future work.

\paragraph{Methodology}
We use the pre-trained \href{https://huggingface.co/microsoft/unixcoder-base}{\texttt{unixcoder-base}} model of \citet{guo2022unixcoder}.
It was pre-trained on 6 programming languages on a variety of tasks, including method naming, on the CodeSearchNet corpus~\citep{husain2019codesearchnet}.
We chose this model since it was the only publicly available transformer-based model pre-trained for function name prediction at the time of writing.
We fine-tune it on the function naming task on the CodeSearchNet corpus.
To retrieve the set $T$ of concrete suggestions, we use standard beam search with 100 beams.
We lowercase and deterministically subtokenise its predictions on \texttt{camelCase} and \texttt{pascal\_case},
then input the result to \methodName (\autoref{sec:method}).
Of course, any machine learning model that generates sequences could be used here as the upstream models, but since \methodName and the baselines use the same outputs of the upstream model, the relative performance difference will remain.
Hence, we evaluate using just the model of \citet{guo2022unixcoder} for simplicity.

\paragraph{Dataset and Baseline}
To collect a Python testset that minimises leakage, we look into the repositories in CodeSearchNet and find all files that were fully touched \emph{after} its publication date (20 Sep. 2019) based on git blame, following the advice of \citet{allamanis2019adverse,nie2022impact}.
This provides sufficient assurance that the testset has not been seen during training.
We use the CodeSearchNet code for filtering and extracting functions and their names.
This process yields 250k test samples.

As a baseline, we use the standard method for improving precision at the cost of recall: if the top suggestion in $T$ has a probability less than a threshold, the wildcard prediction ``$\star$'' is returned.
If the top suggestion has a probability above the threshold, then that single
suggestion, which cannot contain a wildcard, is returned.

\begin{figure}[t]
{\small
\begin{align*}
  \textsc{RegexAcc}\left(\{s^*\}, t\right) = \frac{\textsc{Sub}(\{s^*\})}{\textsc{Sub}(t)}\prod_{s\in\{s^*\}}\mathbb{I}(m(s,t)) \\
  \textsc{CM}\left(\{s^*\}, t\right) = \mathbb{I}(\vert\textsc{Sub}(\{s^*\})\vert > 0)\prod_{s\in\{s^*\}}\mathbb{I}(m(s,t))
\end{align*}
\vspace*{-2mm}
}%
\caption{Evaluation Metrics Definition: RegexAcc and CM (Complete Matches). Here, 
$\textsc{Sub}(\cdot)$ deterministically subtokenises a set of patterns removing any wildcards, $\{s^*\}$ are the suggested patterns, and $t$ is the ground-truth name.
}
\label{fig:metrics}
\vspace*{-6mm}
\end{figure}

\paragraph{Metrics} 
We want to measure the ability of a pattern $s$ to predict at least some
subtokens of a ground-truth name $t$. We deem a pattern useful if
$s\neq [\star]$ and $m(s,t)$ holds, \ie the pattern is non-trivial and it
matches the ground truth. To measure performance, we employ two metrics:
RegexAcc and complete match (CM), defined in \autoref{fig:metrics}. We adapted
RegexAcc from \citet{guo2021learning}; it yields zero if any of the predicted
patterns in $\{s^*\}$ does \emph{not} match the ground-truth; otherwise, it
returns the percent of subtokens (correctly) recalled. In \autoref{fig:lattice
sample}, the RegexAcc, given the ground-truth name
\texttt{load\_all\_msgpack\_l\_gz}, of
``\texttt{load}$\star$\texttt{msgpack}$\star$'' is $\frac{2}{5}=0.4$, but is 0 for
``\texttt{load}$\star$\texttt{messages}$\star$''.  We use RegexAcc
to measure the utility of our abstract solutions for the naming task, reasoning
that matched subtokens will help a developer think of the correct name. False
alarms are the bane of anomaly detection.  Thus, CM first drops any pattern
set that trivially matches because it contains $\star$ by itself, then requires
every remaining nontrivial pattern to match.  In short, it measures how often a
pattern set $\{s^*\}$ would report an accurate warning on a token $t$.  

\subsection{Results}

\begin{figure}[t]
  \begin{subfigure}[b]{\columnwidth}
    \centering
    \includegraphics[width=\textwidth]{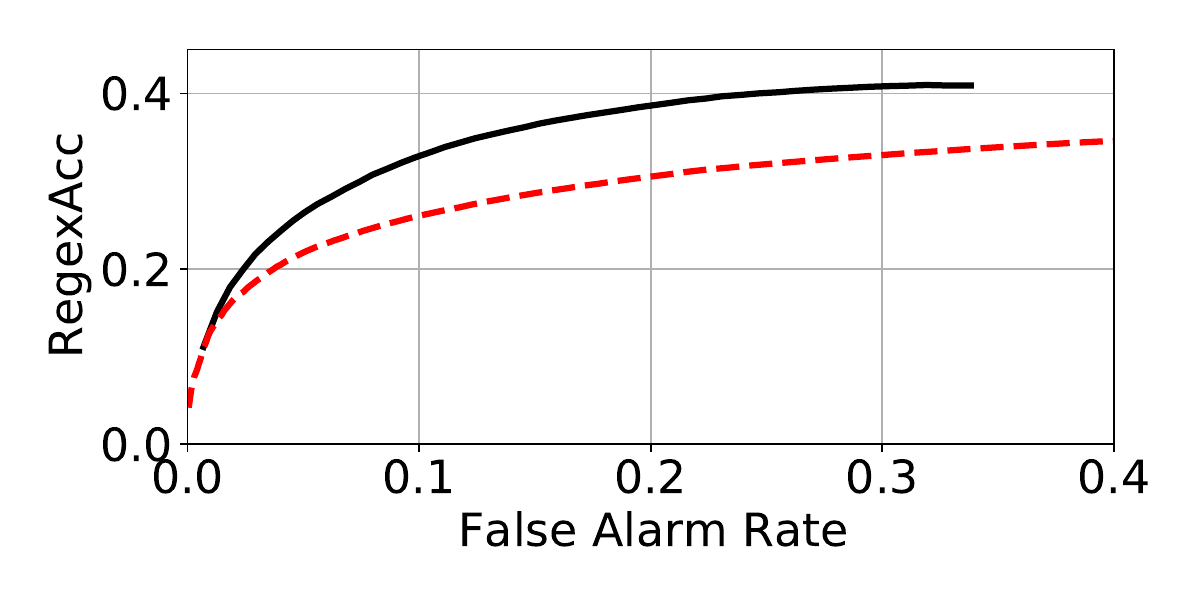}
    \caption{Naming Task:  False Alarm Rate \vs RegexAcc.}\label{fig:fpr vs regex}
  \end{subfigure}
  \begin{subfigure}[b]{\columnwidth}
    \centering
    \includegraphics[width=\textwidth]{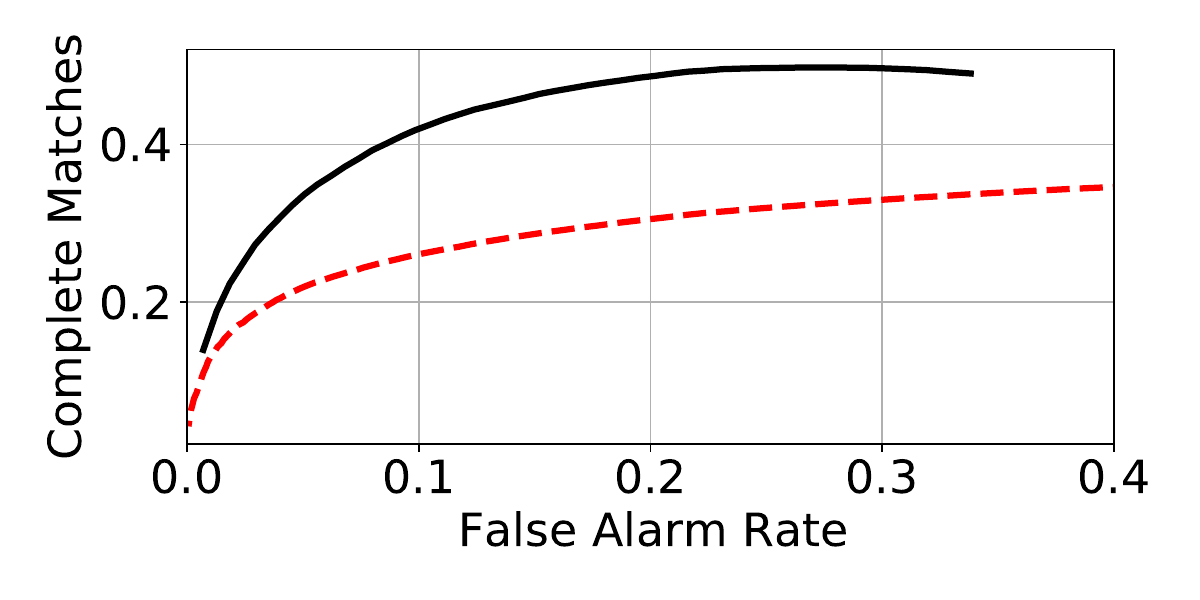}
    \caption{Anomaly Task: False Alarm Rate \vs Complete Matches.}\label{fig:fpr vs nme}
  \end{subfigure}
	\caption{\methodName (black solid line) \vs standard filtering of top suggestion (red dashed line). 
  Since $\theta>0.5$, \methodName can only be used for low false alarm rates.}
\vspace*{-6mm}
\end{figure}

\autoref{fig:fpr vs regex} and \autoref{fig:fpr vs nme} compare \methodName  (black solid line) to the common probability-based filtering baseline (red dashed line).
The results show that for a given false alarm rate (FAR), \ie frequency of suggestions that do \emph{not} match the ground-truth. \methodName achieves better RegexAcc and more complete matches.
For example, for a false alarm rate of 10\% \methodName achieves 33\% RegexAcc with 42\% complete matches, while a standard suggestion filtering strategy achieves only 26\% RegexAcc with 26\% complete matches, \ie
the 61\% improvement reported in the abstract ($\frac{0.4}{0.26}$) at $\text{FAR} = 0.1$.
This suggests that \methodName can provide valuable, precise, albeit partial, naming suggestions better than the baseline and that it is well-suited for anomaly detection since it can better capture (non-trivial) patterns while raising fewer false alarms.

\section{Discussion and Conclusion}
We presented \methodName, a simple method for distilling a set of predictions into a lattice of abstract patterns.
While we showed that this lattice can be used for naming suggestion and anomaly detection, \methodName can also summarise the output of any sequence model in an interpretable way.

\bibliography{bibliography}

\newpage
\appendix

\section{Lattice Example}
\label{sec:lattice example}
\begin{figure*}
  \includegraphics[width=\textwidth]{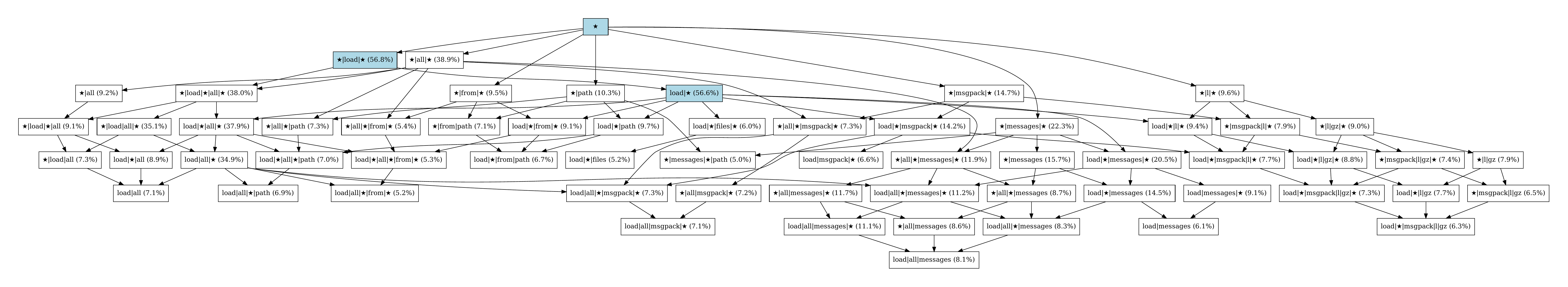}
  \caption{A lattice for the predictions for a Python function with ground-truth name \texttt{load\_all\_msgpack\_l\_gz}.
  Each node contains the pattern $s$ with its tokens separated by ``|''.
  Only the nodes with probability more than 5\% are shown for visual clarity.
  This is a more detailed version of \autoref{fig:lattice sample}.
  The nodes above a threshold $\theta$ of 55\% are shaded blue.
  Best viewed in screen.
  }\label{fig:lattice sample big}
\end{figure*}

\autoref{fig:lattice sample big} shows a more detailed lattice of the output shown in \autoref{fig:lattice sample}.

\end{document}